\documentclass[10pt,journal,compsoc]{IEEEtran}

\ifCLASSOPTIONcompsoc
  \usepackage[nocompress]{cite}
\else
  \usepackage{cite}
\fi
\ifCLASSINFOpdf
\else
\fi
\usepackage{cuted}
\usepackage{hyperref}
\usepackage{url}
\usepackage{multirow}
\usepackage{makecell}
\usepackage[utf8]{inputenc} 
\usepackage[T1]{fontenc}    
\usepackage{hyperref}       
\usepackage{url}            
\usepackage{booktabs}       
\usepackage{amsfonts}       
\usepackage{nicefrac}       
\usepackage{microtype}      
\usepackage{xcolor}         
\usepackage{microtype}
\usepackage{graphicx}
\usepackage{subfigure}
\usepackage{amsmath}
\usepackage{amssymb}
\usepackage{mathtools}
\usepackage{amsthm}
\usepackage{subfigure}
\usepackage{rotating}
\usepackage[linesnumbered, ruled]{algorithm2e}

\SetKwRepeat{Do}{do}{while}

\begin{document}
%
\title{Multi-scale Attention Flow for Probabilistic Time Series Forecasting}
%
%
%
%

\author{Shibo~Feng, Chunyan~Miao, \textit{Fellow}, \textit{IEEE}, Ke~Xu, Jiaxiang~Wu, Pengcheng~Wu, Yang~Zhang, Peilin~Zhao
\IEEEcompsocitemizethanks{\IEEEcompsocthanksitem Chunyan Miao, Shibo Feng and Pengcheng Wu are with the LILY Research Center, Nanyang Technological University, Singapore, 639798. \protect E-mail: ascymiao@ntu.edu.sg; shibo001@ntu.edu.sg; pengcheng.wu@ntu.edu.sg.

\IEEEcompsocthanksitem Ke Xu, Jiaxiang Wu, and Peilin Zhao are with Tencent AI Lab, Shenzhen, China, 518057. \protect Email: kaylakxu@tencent.com; jonathanwu@tencent.com; masonzhao@tencent.com.
\IEEEcompsocthanksitem Yang Zhang is with Department of Radiology, Qilu Hospital of Shandong University. \protect drzhy001@163.com.
\IEEEcompsocthanksitem Shibo Feng and Ke Xu are the co-first authors.
\IEEEcompsocthanksitem Peilin Zhao and Yang Zhang are the corresponding authors.
}

\thanks{This work is done when Shibo Feng works as an intern in Tencent AI Lab.}}

\IEEEtitleabstractindextext{%
\begin{abstract}
The probability prediction of multivariate time series is a notoriously challenging but practical task. On the one hand, the challenge is how to effectively capture the cross-series correlations between interacting time series, to achieve accurate distribution modeling. On the other hand, we should consider how to capture the contextual information within time series more accurately to model multivariate temporal dynamics of time series. In this work, we proposed a novel non-autoregressive deep learning model, called Multi-scale Attention Normalizing Flow(MANF), where we combine multi-scale attention with relative position information and the multivariate data distribution is represented by the conditioned normalizing ﬂow. Additionally, compared with autoregressive modeling methods, our model avoids the influence of cumulative error and does not increase the time complexity. Extensive experiments demonstrate that our model achieves state-of-the-art performance on many popular multivariate datasets.
\end{abstract}

}

\maketitle

\IEEEdisplaynontitleabstractindextext

%
\IEEEpeerreviewmaketitle

\IEEEraisesectionheading{\section{Introduction}\label{sec:introduction}}

%
%
%
%
\IEEEPARstart{F}{orecasting} the future movement and value of time series is a key component of formulating effective strategies in most industrial fields. Specific applications include: forecasting volatility and trends of the financial markets \cite{sezer2020financial}, energy and electricity demand \cite{cao2021spectral}, traffic flow \cite{lu2021temporal}, the transmission of COVID-19 \cite{zeroual2020deep}, recognition of human and videos \cite{zhang2019making, luo2017adaptive, chen2019semisupervised}, and so on. Compared with univariate prediction tasks, multivariate time series forecasting is more common in practical industry. For instance, the electricity consumption of multiple clients, which is tracked by the local power company -- amounting to billions of power time series.

Most of the statistical time series forecasting methods are dependent upon univariate forecasting models. Such methods need expert knowledge and feature engineering as the premise to describe the potential periodicity and seasonality of the time series sequences, which determines that these models contain inherent defects of high cost and strong stochasticity. As described in \cite{hyndman2018forecasting}, the main form of univariate models can be described as linear-based parametric statistical models such as AR \cite{gneiting2014probabilistic}, ARIMA \cite{box1970distribution}, ARIMAX \cite{williams2001multivariate} and Exponential smoothing (ES) \cite{gardner2006exponential}. The limited expression ability of these statistical forecasting methods makes researchers pay more attention to the methods of RNN-family like LSTM \cite{hochreiter1997long}. Recently, the self-attention mechanism has made remarkable achievements in various fields, and therefore, researchers introduce Transformer-based \cite{vaswani2017attention} models to make up for the shortboard of the RNN-family in long-term modeling. However, these approaches are to abandon multi-variate forecasting entirely and perform univariate forecasting (i.e., fit a separate model per series). In most realistic prediction scenarios, cross-series correlations/effects are the key to complete precise prediction. For example, in the financial market, there exists a strong correlation between stocks in the same and complementary sectors\cite{cao2020spectral}, \cite{sezer2020financial}. In the energy field, the spatial correlation effects between energy time series will affect the effective allocation of resources\cite{wang2019review}, \cite{khan2020relationship}. Also, in the semantic segmentation tasks, CTNet \cite{li2021ctnet} uncovered the spatial contextual dependency between pixels by exploring the correlation between pixels and categories. Furthermore, these approaches cannot leverage the extra information provided from related series in case of noise or sparsity. E.g., sales are often sparse (e.g., one sale a month for a particular product and store), so the sales rate cannot be accurately estimated from a single series.

To match the multivariate time series settings, one common approach is to fit a single multi-output model to predict all series simultaneously. These models include GARCH\cite{van2002go}, VAR-Lasso\cite{tibshirani1996regression}\cite{lutkepohl2013vector}, temporal convolutional neural networks (TCNs)\cite{wan2019multivariate} and some combination of RNN, CNN-family \cite{cheng2020towards, zhou2020parallel, rodrigues2020beyond}. However, these multivariate models often assume a simple parametric distribution and do not scale to high dimensions, which could cause model overfitting and reduce their generalization ability.

Recently, the attention mechanism has shown excellent performance in the Natural Language Processing field. Within the setting of multivariate time series forecasting, Shih\cite{shih2019temporal} proposed filters-based attention to extracting time-invariant temporal patterns and used its frequency domain information for multivariate forecasting. Du et al. \cite{du2020multivariate} proposed a temporal attention encoder-decoder model, leveraging Bi-LSTM layers with a temporal attention mechanism to study long-term dependency and hidden correlation features. These attention-based encoder structures always follow the same process in that they decomposed the sequence features from different perspectives and decoupling learning temporal correlation features through deep learning methods. However, the task of time series prediction is more inclined to achieve extrapolation. If we combine the sequence model with the generative model of effective modeling data distribution, it may 
complete the extrapolation task of multivariate time series forecasting. 
.
In general, these state-space models effectively integrate exogenous covariates and realize end-to-end training that makes them popular among researchers.
However, vanilla attention has a number of limitations in tackling multivariate time series forecasting: (i) the global self-attention mechanism is insensitive to the local context, whose dependencies sometimes may play a key role in forecasting; (ii) The point-wise dot-product self-attention mechanism lacks the capability of accurately learning of position and timestamp features since they integrate the two parts to perform the calculation. Moreover, the distribution of attention weight does not consider the factors of distance and sequence, and it usually has a long-tailed distribution, that is, the multi-layer attention mechanism only repeatedly focuses on the most important information while ignoring the more important and relatively important information. In this work, we propose relative location information and a multi-scale mechanism to make up for the shortcomings of vanilla attention that cannot learn position factors and hierarchical information of time context.

To model the full predictive probability distribution, methods typically resort to tractable distribution classes or some type of low-rank approximations, regardless of the true data distribution. To model the distribution in a general fashion, one needs probabilistic methods with tractable likelihoods. As a type of powerful generative model, normalizing ﬂow based structures have a simple density through a series of transformations to produce a richer and potentially more multi-modal distribution, which is suitable for modeling high-dimensional multivariate time series. Compared with approximate models like variational autoencoders \cite{kingma2019introduction} and generative adversarial nets \cite{goodfellow2014generative}, normalizing ﬂows construct a reversible bijective mapping function.

So far, more sophisticated time series decomposition models have shown strong competitiveness in time series tasks, for example, \cite{ding2020hierarchical, xu2021autoformer, hyndman2018forecasting}. In general, there are two strategies to model hierarchical information in time series: preprocessing \cite{taylor2018forecasting, oreshkin2019n} before modeling and learning through the neural networks \cite{xu2021autoformer}. Since the complex correlations between multivariate variables in time series is difficult to decompose explicitly, and the interaction between timestamps at different distances is difficult to be calculated quantitatively, our work combines relative position information and multi-scale attention to learn hierarchical time-series features.

In this work, we proposed an end-to-end trainable non-autoregressive architecture for probabilistic forecasting that explicitly models hierarchical temporal patterns of time series with novel multi-scale attention. And on this basis, a conditional normalizing flow RealNVP \cite{dinh2016density} is introduced to model multivariate time series. Experiments show that our model can effectively capture the accurate dependency structure and we establish a new state-of-the-art result on over competitive baselines on diverse real-world datasets. Additionally, compared with the autoregressive modeling methods, we avoid the cumulative error of multi-step forecasting and improve the parallelism in a non-autoregressive way. In summary, the main contributions of our work include:
\begin{itemize}
\item We propose MANF that can capture the multi-scale temporal dependency and dynamic multivariate time series correlations simultaneously. To our knowledge, this is the first work to introduce the combination of the sequence model and generative model in a non-autoregressive way for multivariate time series forecasting.

\item We propose a novel multi-scale attention mechanism with dynamic relative position embedding to model local context information and global long-term patterns and introduce normalizing flow structure to model high-dimensional multivariate time series without increasing the time complexity in a non-autoregressive way.

\item We perform extensive experiments with multiple multivariate forecasting datasets, demonstrating superior performance compared with the recent state-of-the-art forecast methods, for probabilistic predictions.
\end{itemize}

\section{Related Work}
\label{Related}
There are two kinds of work that are more relevant to ours (i) multivariate time series forecasting using deep neural networks; (ii) probabilistic modeling, prediction, and forecasting for high-dimension multivariate time series. Recently, deep neural network-based models have shown strong competitiveness in multivariate time-series prediction. Li et al.\cite{li2021learning} proposed the deep state space model, leveraging parameterized networks to construct the nonlinear emission model and transition model for multivariate forecasting, LSTNet\cite{lai2018modeling} leveraged correlations between multiple time-series through a combination of 2D convolution and recurrent structures. DeepGLO \cite{sen2019think} combined the global matrix factorization model regularized by TCN\cite{lea2017temporal} and local temporal network for time series forecasting. N-BEATS\cite{oreshkin2019n} proposed a deep neural architecture based on backward and forward residual links and a very deep stack of fully-connected layers for univariate times series point forecasting. DSMHN \cite{jin2020deep} explored the inter-modality correlation structure and the intra-modality semantic label information for scalable image-text and video-text retrievals. However, these deterministic models can neither model the uncertainty of the required forecasting nor fulfill the probabilistic forecasting.

Recently, powerful multivariate forecasting models that are capable of providing uncertainty characterization have received a great deal of attention. DeepState \cite{rangapuram2018deep} encoded time-varying parameters for linear state space models and performed inference via the Kalman filtering equations. CATN \cite{he2022catn} proposed an end-to-end model of cross attentive tree-aware network to jointly capture the inter-series correlation and intra-series temporal patterns for time series forecasting. STID \cite{shao2022spatial} proposed a simple yet effective baseline for MTS forecasting by attaching spatial and temporal identity information. However, our proposed method is different from these works in state modeling. We avoid the need for some approximations required in a variational inference framework.
To our knowledge, most existing works rely on the autoregressive model, for instance, Transformation autoregressive networks \cite{oliva2018transformation} model the multivariate variable as conditional distribution, where the conditioning is derived from the state of RNN, and then transformed via a bijection. PixelSNAIL \cite{chen2018pixelsnail} substituted RNN units with causal convolutions and self-attention mechanisms to study long-term temporal dependencies and then complete the density estimation. Based on the sequential modeling structure like LSTM, Transformer, Kashif, et al. \cite{rasul2020multivariate} introduced conditional affine coupling layers to model the output variability. Compared with the above autoregressive models, our work firstly enjoys an obvious advantage in the generation speed and avoids the introduction of cumulative error, which can improve the accuracy of prediction distribution.

As a non-autoregressive multivariate forecasting work, Normalizing Kalman filters (NKF) \cite{de2020normalizing} introduced the normalizing flow as an enhanced component for the classical linear Gaussian state-space model (LGM) and achieve exact inference through ﬁltering, smoothing and likelihood computation. However, LGM-based models have inherent defects, within some complex environments, it is quite fragile to depict the non-stationary and high stochasticity of multivariate data with LGM. Since it is a competitive non-autoregressive sequence modeling structure, we will compare our model to this method in what follows.

\begin{figure*}[htp]
  \begin{center}
  \includegraphics[width=1.70\columnwidth]{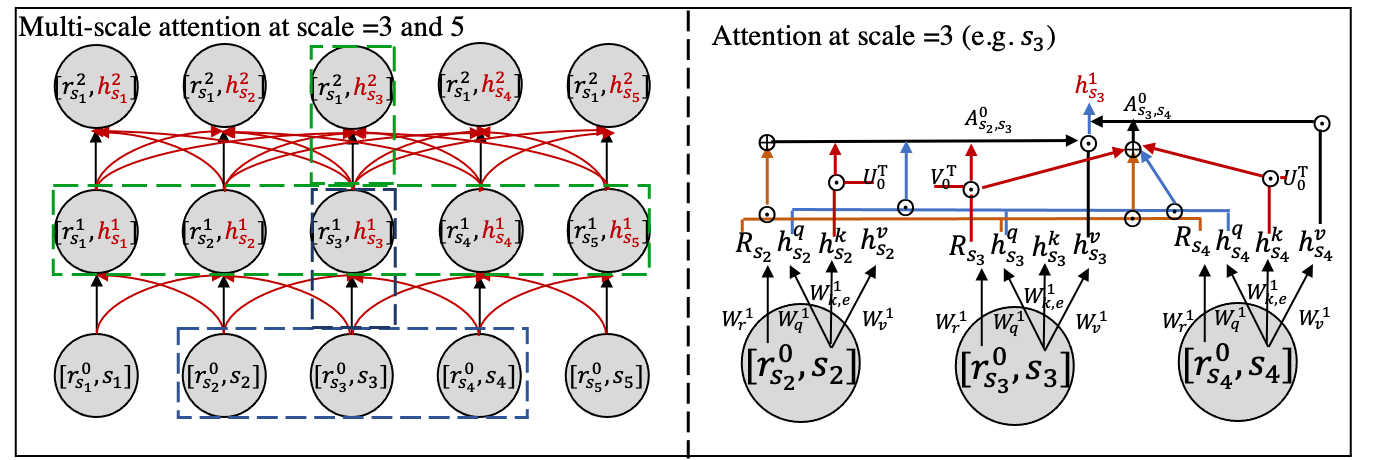}
  \caption{Multi-scale attention mechanism with size =3,5. $W_{r}^{i}$ is the learnable parameter of the relative position $R_{s_{i}}$ of the i-th layer}
  \label{Fig2}
  \end{center}
\end{figure*}

\section{Preliminaries}

\subsection{Normalizing Flows for Probabilistic Modelling}

The basic principles behind ﬂow-based generative models were ﬁrst described by \cite{deco1995higher}. Dinh et al.re-discovered and refined the flow-based model within the modern context such as Non-linear Independent Component Estimation(NICE) \cite{dinh2014nice}, which stacked multiple additive coupling layers to endow the model with strong fitting ability and unit Jacobian determinant at the same time. RealNVP further combined the additive and multiplicative coupling layers into a general ``affine coupling layer" and introduced convolution operation to deal with images more effectively and reduce the number of parameters.

The general process of applying normalizing flows for probabilistic modeling is shown in \cite{papamakarios2021normalizing}. Given input $X=\mathbb{R}^{D}$ whose densities are represented as $p_{x}$ are transformed into a simple and tractable density such as a spherical multivariate Gaussian distribution $p_{z}$ on the space $Z=\mathbb{R}^{D}$. Additionally, the mapping function $\mathcal{F}: X \rightarrow Z$ is composed of a sequence of invertible transformations. Thus, $p_{x}$ can be described as 
\begin{equation}
\label{base}
p_{x}=p_{z}(Z)\left|\operatorname{det}\left(\frac{\partial \mathcal{F}(x)}{\partial x}\right)\right|,
\end{equation}
where $\frac{\partial \mathcal{F}(x)}{\partial x}$ is the Jacobian of $\mathcal{F}$ at $x$. Normalizing ﬂows have the property that the inverse $\mathcal{F}^{-1}$ should be easy to evaluate and compute the Jacobian determinant concerning the parameters of the flows $\mathcal{F}$.

The bijection introduced by RealNVP \cite{dinh2016density} of particular interest for our work, is called the affine coupling layer whose determinant can be easily computed with its diagonal terms of a triangular matrix. Specifically, it leaves the part of its inputs unchanged and transforms the other part via functions of the un-transformed variables. Given input $X^{D}$ and $d<D$, the transformation is exploited by deﬁning the output of an afﬁne coupling layer as follows:

\begin{equation}
\begin{gathered}
m^{1: d}=x^{1: d},  \\
m^{d+1: D}=x^{d+1: D} \odot \exp \left(s\left(x^{1: d}\right)\right)+t\left(x^{1: d}\right),
\end{gathered}
\end{equation}


where $s$ and $t$ are arbitrarily complex scale and translation functions from $\mathbb{R}^{d} \mapsto \mathbb{R}^{D-d}$, parameterized by the weights of neural networks $\theta$ and $\odot$ is the element-wise or Hadamard products. Generally, in order to improve the nonlinear expression ability of $\mathcal{F}$, a number of coupling layers are composed together to construct a normalizing flow, this is, $X \mapsto \mathcal{M}_{1} \mapsto \cdots \mapsto \mathcal{M}_{N-1} \mapsto \mathcal{Z}$. Under the change of variables of Eq.~\ref{base}, the probability density function $\log p_{x}(X)$ of the model given an input $x$ can be written as:
\begin{equation}
\begin{gathered}
\log p_{x}(X) = \log p_{z}(Z)+\sum_{i=1}^{N} \log \left|\operatorname{det}\left(\partial m_{i} / \partial m_{i-1}\right)\right|, \\
\log \left|\operatorname{det}\left(\partial m_{i} / \partial m_{i-1}\right)\right|=\log \left|\exp \left(\operatorname{sum}\left(s_{i}\left(m_{i-1}^{1: d}\right)\right)\right)\right|,
\end{gathered}
\end{equation}
This model is parameterized by the weights of the $s$ and $t$ neural networks $\theta$, we maximize the average log-likelihood in each batch $\mathcal{D}$ given by
\begin{equation}
\mathcal{L(\theta)}=\frac{1}{|\mathcal{D}|} \sum_{\mathbf{x} \in \mathcal{D}} \log p_{\mathcal{X}}(\mathbf{x} ; \theta)
\label{loss}
\end{equation}

In practice, we introduce batch normalization(BN) \cite{ioffe2015batch} as a bijection, which is embedded in the training and inference process of the normalizing flows, so as to ensure the robustness of network structure. The specific operation is described as follows: 
\begin{equation}
\begin{gathered}
\hat{x} \longleftrightarrow \frac{x-\bar{u}}{\sqrt{\bar{\sigma}^{2}+\epsilon}}, \\
y \longleftrightarrow \gamma \hat{x}+\beta \equiv B N_{\gamma, \beta}(\hat{x}),
\end{gathered}
\end{equation}
where $\gamma$, $\beta$ both are the learnable parameters, $\bar{u}$, $\bar{\sigma}^{2}$ are the mini-batch mean and variance respectively. $\epsilon$ is a constant added to the mini-batch variance for numerical stability.

\subsection{Multi-Scale Attention Network}

Recently, models based on self-attention mechanisms like Transformer \cite{vaswani2017attention} and BERT \cite{devlin2018bert} achieve many exciting results in sequence modeling. However, self-attention modules inherently lack appropriate inductive bias, so the trained models have limited generalization ability. Specifically, the distribution of attention weight does not consider the factors of distance and sequence, and its distribution has the problem of sparsity, that is, the multi-layer attention mechanism only repeatedly focuses on the most important information while ignoring the more important and relatively important information.

In order to capture the local bias and encourage the combined features to be outstanding and sharp, we set the small-scale attention module into the shallow layer in the Transformer encoder. With the depth of the model, we establish a wider scale of attention, to make our model form a gradual process of inductive learning. Moreover, different scales of attention will result in different position information represented by the same timestamp at different depths of the model, where we need to introduce relative position embeddings into multi-scale attention that can refer to Fig.~\ref{Fig2}. Concretely, given a sequence of vectors $S=\left(s_{1}, \ldots, s_{N}\right) \in \mathbb{R}^{N\times D}$ with length $N$. $\Theta$ is a set of parameters where the scales are the constant numbers or the ratios according to the sequence length. Practically, one can create a set of relative position encodings $ \mathbb{R}^{N\times D}$, where the $i$-th row $R_{i}$ indicates 
the relative position of $i$. The multi-scale attention $\operatorname{A}$ at timestamp $i$ within scale $\Theta_{l}$ is defined as:
\begin{equation}
\label{A}
\begin{gathered}
\mathrm{A}\left(s_{i}, \Theta_{l}\right)=\sum_{h=1}^{H} \mathcal{G}^{h}\left(Q_{h}\left(s_{i}\right), \mathcal{W}_{i}\left(K_{h}, \Theta_{l}\right)^{T}\right) \cdot \mathcal{W}_{i}\left(V_{h}, \Theta_{l}\right), \\
\mathcal{G}^{h}(x, y)= (x + u_{\Theta_{l}}^{T}) W_{k, \Theta_{l}}^{h} y + (x + v_{\Theta_{l}}^{T}) W_{k, \Theta_{l}}^{h} R_{x-y}^{\Theta_{l}}, \\
Q_{h}=S W_{h}^{Q}, K_{h}=S W_{h}^{K}, V_{h}=S W_{h}^{V}, \\
\mathcal{W}_{i}\left(x, \Theta_{l}\right)=\left[x_{i-\Theta_{l}}, \ldots, x_{i+\Theta_{l}}\right], \\
\end{gathered}
\end{equation}
where $Q_{h}$, $K_{h}$:$R^{D} \rightarrow R^{m}$ are query and key functions respectively. $V_{h}$: $R^{D} \rightarrow R^{D}$is a value function, $H$ denotes the number of heads. $\Theta_{l}$ represents $l^{th}$ window size in set $\Theta$, which is corresponding to the $l^{th}$ encoder layer in transformer. Additionally, $W_{k, \Theta_{l}}^{h}$ and $R_{x-y}^{\Theta_{l}}$ denote the location-based weighted key vectors and relative counterpart position in the scale $\Theta_{l}$. Notably, different from \cite{dai2019transformer}, we set different learnable variables $u_{\Theta_{l}}^{T}$ and $v_{\Theta_{l}}^{T}$ at different scales, and therefore, the attention calculation at the same timestamp under different scales can learn accurate timing information. \textbf{Multi-Scale Transformer layer} of $H_{l}$ is defined as:
\begin{equation}
\begin{gathered}
O_{l+1}=\text {LayerNorm}\left(H_{l}+\operatorname{ReLU}\left(\operatorname{A}\left(H_{l}, \Theta_{l}\right)\right)\right),\\
H_{l+1}= \text {Positionwise-Feed-Forward}(O_{l+1}),\\
\text{LayerNorm}(x)=x-E[x] / \sqrt{\operatorname{Var}[x]+\epsilon} \cdot \gamma+\beta,  \\
\text{FFN}(x)=\max \left(0, x W_1+b_1\right) W_2+b_2,
\end{gathered}
\end{equation}
where $l$ is the layer index. And each of the layers in our encoder and decoder contains a fully connected feed-forward network (FFN), which is applied to each position separately and identically. This consists of two linear transformations with a ReLU activation in between.

\section{Multi-scale Attention Flows}

\begin{figure*}[t]
\begin{center}
  \includegraphics[width=1.70\columnwidth]{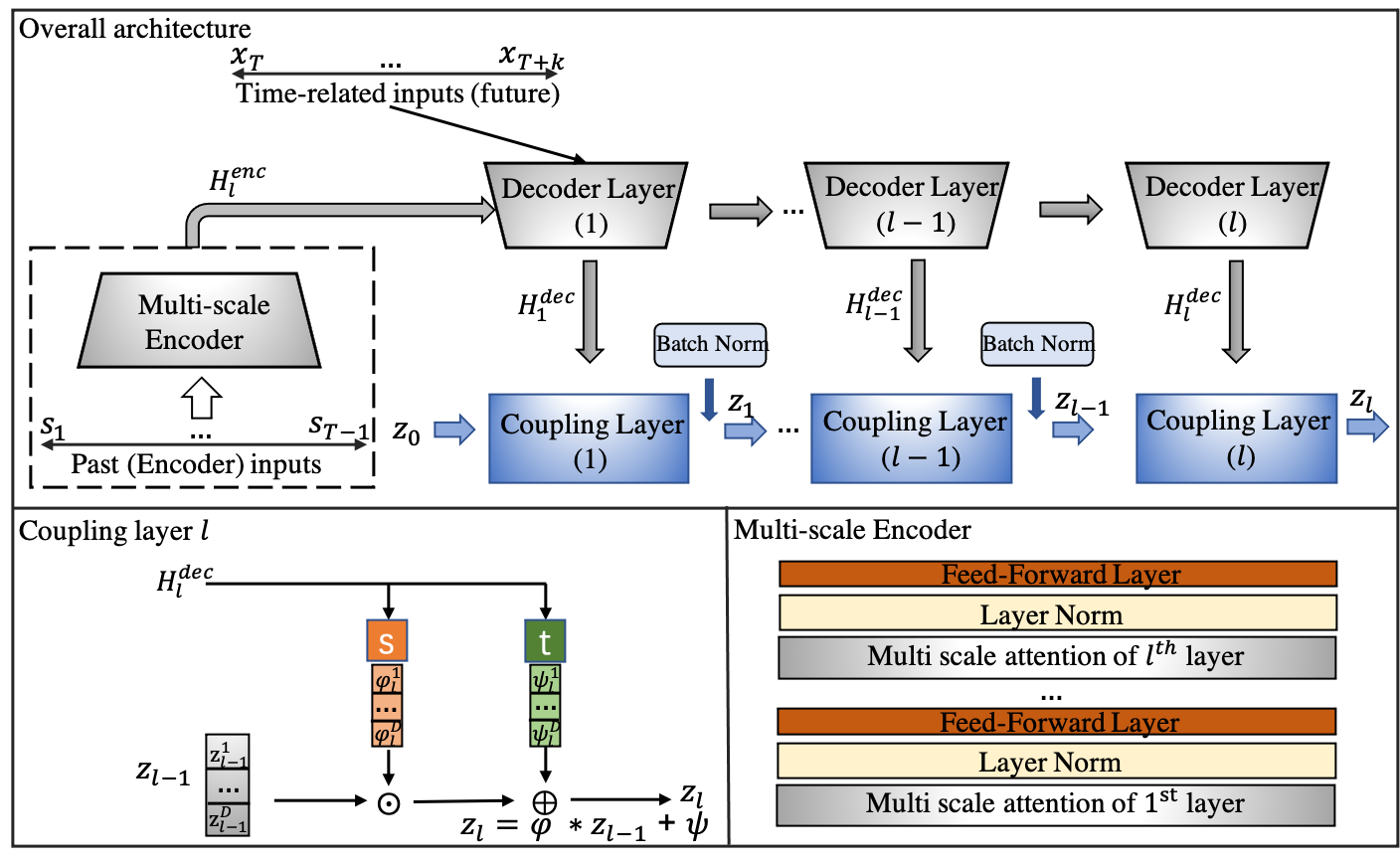}
  \caption{MANF-structure: a multi-attention based conditional normalizing flow model. $H_{l}^{enc}$ and $H_{l}^{dec}$ are the output in $l$-th encoder layer and decoder layer, respectively. $s_{1}, s_{T-1}$ is the embedding of target and time-related covariates in the historical period. $x_{T}, x_{T+k}$ are only the time-related covariates in the future. $z_{0}$ is sampled from  $\mathcal{N}(\mathbf{0}, \mathbf{I})$, $z_{l}$ is the output of the generative model.}
  \label{Fig1}
\end{center}
\end{figure*}

We denote the entities of a multivariate time series by $s_{i}$, where $i$ is the time index. Further, let  $x_{i}$ be time-varying covariate vectors associated with each univariate time series at timestamp $i$. We will consider time series with $t \in[1, T+k]$, where for training we split this time series by the past window $[1, T)$ and prediction window $[T, T+k]$. In this section, we describe our model from two aspects: sequence modeling and generative network. Finally, the overall structure of our work is shown in Fig.~\ref{Fig1} and specific algorithm can refer to algorithm ~\ref{algo}.

\begin{algorithm} \small  \KwData{The scale set $\Theta$, history input $s_{1}, s_{T-1}$,  future time-related covariates $x_{T}, x_{T+k}$, position embedding $PE_{T, T+k}$, gaussian sample $z_{0}$, layernum $l$ and learnable parameters $u_{\Theta}^{T}$,$v_{\Theta}^{T}$.} \KwResult{$z_{out}$ // this is target sequence}  Generation\; $H_{1}\leftarrow A\left(s_{1, T-1}, \Theta_1\right)$; // eq ~\ref{A}, eq ~\ref{need} \\
\For{$i\leftarrow 2$ \KwTo $l$}{$O_{i+1}\leftarrow \text {Layernorm}\left(H_{i}+\operatorname{ReLU}\left(\operatorname{A}\left(H_{i}, \Theta_{i}\right)\right)\right)$; \\
$H_{i+1} \leftarrow \text {Positionwise-Feed-Forward}(O_{i+1})$;}
$H_{1}^{dec}\leftarrow \text {Attn}\left(x_{T,T+k}, H_{l}, \operatorname{PE}_{T,T+k}\right)$; // eq ~\ref{flow} \\
$z_{1}\leftarrow z_{0} \odot \exp \left(s\left(H_{1}^{dec}\right)\right)+t\left(H_{1}^{dec}\right)$; \\
\For{$i\leftarrow 2$ \KwTo $l$}{$O_{i+1}^{dec}\leftarrow \text {Layernorm}(H_{i}^{dec}+\operatorname{ReLU}\left(\operatorname{Attn}\left(H_{i}^{dec}, H_{l},)\right)\right)$; \\
$H_{i+1}^{dec} \leftarrow \text {Positionwise-Feed-Forward}(O_{i+1})$;\\
$z_{i+1}\leftarrow z_{i} \odot \exp \left(s\left(H_{i}^{dec}\right)\right)+t\left(H_{i}^{dec}\right)$; // eq ~\ref{flow}} \emph{return $z_{l}$ as the $z_{out}$}
\caption{Multi-scale Attention Flow (MANF)} \label{algo}\end{algorithm}

\textbf{Sequence modeling} For modeling the time evolution, our work combines learnable relative position information with multi-scale attention. The multi-scale attention mechanism can be interpreted as: (1) \emph{Locality Perception}: the small-scale attention is sensitive to the local context, whose dependencies are much more important in multivariate time series. (2) \emph{Long-term Hierarchy Poverty}: the combination of hierarchical scale attention is the capability of utilizing the hierarchical structure of time series (e.g. intra-day, intra-week, and intra-month features in time series independently. Stacking different scales of attention layers to make the model build on the knowledge of local contextual content to capture high-order global information.

In the encoder, we increase the scale with the deepening of encoder layers. Notably, we modify the relative position embedding of each timestamp in the past window according to different scales. The size of the scale increases with the depth of the encoder layer and $l$-th scale attention mechanism $\operatorname{A}_{\Theta_{l}}$ performed on hidden state $H_{i}^{enc}$ is described as follows:
\begin{equation}
\begin{aligned}
&H_{i+1}^{enc}=A\left(H_{i}^{enc}, \Theta_{i+1}\right) \\
&H_{1}^{enc}=A\left(S_{1, T-1}, \Theta_1\right),
\end{aligned}
\label{need}
\end{equation}
where $\operatorname{A}$ is depicted in Eq.~\ref{A}, $H_{i}^{enc}$ is the outputs of the $i$-th layer and $i \in \text { for } i=1, \ldots l$. Besides, $\Theta_{i}$ represents the $i^{th}$ scale in the set $\Theta$ and $S_{1,T-1}$ is the embedding of input items within time $[1, T-1]$, respectively.\\
\textbf{Generative network} Compared with the previous autoregressive approach \cite{rasul2020multivariate}, our generative procedure is a non-autoregressive structure in the sense that observations $s_{t}$ in the prediction window are never fed to the model. Instead, we directly take the outputs of the decoder layers as the conditions that become the inputs to the normalizing flows. In the experiment, we found that the non-autoregressive structure avoids the negative effect caused by the increase in prediction length. We take the low-level output of the decoder layer as the "basic knowledge" and deep ones as the "reinforcement knowledge" then the stacked normalizing flows can effectively generalize biases, resulting in a powerful generative model. Given the covariant vectors $X_{T,T+k}$ and  $\operatorname{PE}_{T,T+k}$ (vanilla sinusoidal position encoding \cite{vaswani2017attention}) in prediction window $[T, T+k]$. And the normalizing flow transformation of step $l$ ($z_l$) is defined as:
\begin{equation}
\begin{gathered}
H_{i+1}^{dec}=\text {Attn}\left(H_{i}^{dec}, H_{l}^{enc}, \operatorname{PE}_{T,T+k}\right), \\
H_{1}^{dec}=\text {Attn}\left(X_{T,T+k}, H_{l}^{enc}, \operatorname{PE}_{T,T+k}\right), \\
z_{i+1}=z_{i} \odot \exp \left(s\left(H_{i}^{dec}\right)\right)+t\left(H_{i}^{dec}\right), \\
z_{0} \sim p_{z},
\end{gathered}
\label{flow}
\end{equation}

where $i \in \text { for } i=0, \ldots, l$, Attn is the vanilla self-attention in Transformer, $H_{i}^{dec}$ is the output embedding in $i^{th}$ decoder layer. $p_{z}$ typically is a Gaussian distribution. The covariates $X_{T, T+k}$ in our work are composed of time-dependent (e.g. day of the week, an hour of the day) and time-independent embeddings (e.g. categories).

In general, compared with the autoregressive model commonly used in time series forecasting, we have constructed a new one-shot framework. In the encoder, we leverage the multi-scale attention to study the multi-angle relative information of the historical sequence at different scales. In the decoder, we construct a non-autoregressive Transformer based normalizing flow generation model and complete the effective generation of the target sequence by corresponding to the multi-layer flow structure and Transformer.\\
\textbf{Efficient computation} The combination of multi-scale attention and normalizing flow in our work will not significantly improve the time complexity. The time complexity of a multi-scale Transformer is $\mathcal{O}(R T D)$, and RealNVP is $\mathcal{O}(D)$. Unlike autoregressive models, all computation while training and testing, over the time dimension, happens in parallel. Compared with the combination of the autoregressive Transformer or LSTM and flow $\mathcal{O}\left(D^{2} T\right)$, our structure does not increase the computational complexity too much. R is the scale size, T represents the window size and D is the number of elements.

In real-world scenarios, the magnitudes of different time series can vary drastically. To normalize scales, we divide each time series by its past window means before feeding it into the model. At inference, the distributions are then correspondingly transformed with the same mean values to match the original scale. Results in the appendix show that our model is comparable to other autoregressive models in terms of running time and testing time.

\begin{figure*}[ht]
\begin{center}
   \includegraphics[width=0.75\textwidth]{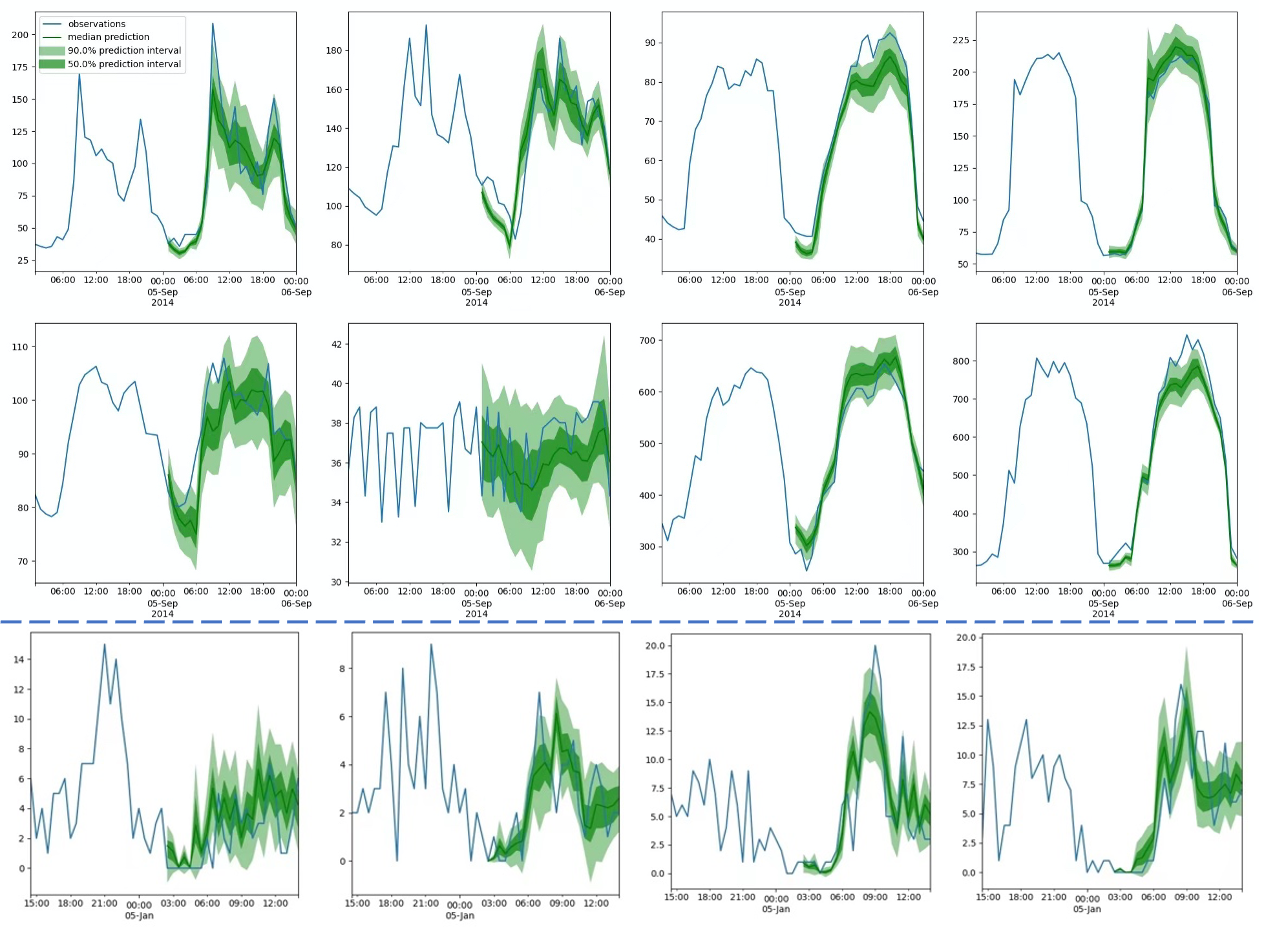}
  \caption{MANF prediction intervals and test set ground-truth for Electricity (above the blue dotted line) and Taxi (below the blue dotted line) data of the 8 of 963 dimensions from a rolling window.}
  \label{okk_2}
\end{center}
\end{figure*}


\section{Quantitative Experiments}

\textbf{Datasets} We extensively evaluate the proposed MANF on six real-world benchmarks, covering the mainstream time series probabilistic forecasting applications, Energy: Solar \cite{lai2018modeling} and Electricity, Traffic: Traffic and Taxi, Economics: Exchange \cite{lai2018modeling} and Wikipedia. During training, we introduced the mixup \cite{zhang2017mixup}, a data augmentation method to ensure the stability of the experiment. The properties of the datasets used in experiments are shown in Table ~\ref{tab:data}\\
\begin{table}[hb]
  \centering
  \caption{Properties of the datasets in experiments}
\resizebox{0.9\linewidth}{!}{\begin{tabular}{cccccc}
\hline
  \textbf{DATASET} & Dimension & Domain & Freq & Total Time Steps & Pred Length  \\ \hline
EXCHANGE & 8  & $\mathbb{R}^{+}$ & Daily & 6,071 & 30 \\ 
SOLAR  & 137  & $\mathbb{R}^{+}$ & Hourly & 7,009 & 24 \\ 
ELECTRICITY  & 370 & $\mathbb{R}^{+}$ & Hourly & 5,790 & 24  \\  
TRAFFIC   & 963 & (0,1) & Hourly & 10,413 & 24 \\  
TAXI    & 1214  & $\mathbb{N}$ & 30-Min & 1,488 & 24 \\ 
WIKIPEDIA   & 2000  & $\mathbb{N}$ & Daily & 792 & 30\\ \hline
\end{tabular}
}
  \label{tab:data}
\end{table}\textbf{Evaluation Metrics} For probabilistic estimates, we report both the continuously ranked probability score across summed time series (CRPS-sum) (\cite{matheson1976scoring, jordan2017evaluating}) and mean square error (MSE) error metrics, to measure the overall joint distribution pattern ﬁt and ﬁt of joint distribution central tendency, respectively. As in \cite{rasul2020multivariate} CRPS measures the compatibility of a cumulative distribution function $P$ with an observation $x$ as
\begin{equation}
\operatorname{CRPS}(\mathcal{F}, x)=\int_{\mathbb{R}}(P(y)-\mathbb{I}\{x \leq y\})^{2} d y,
\end{equation}
where $\mathbb{I}\{x \leq y\}$is the indicator function which is one if $x \leq y$ and zero otherwise. Since CRPS is a widely accepted scoring function, we can employ the empirical CDF of $P$, i.e., $\hat{P}(y)=\frac{1}{N} \sum_{i=1}^{N} \mathbb{I}\left\{X_{i} \leq y\right\}$ with n samples $X_{i} \sim P$ as the approximation of the predictive CDF. CRPS can be directly computed from samples of the conditional distribution at each time point. In practice, we utilize 100 samples to estimate the empirical CDF, and take the CRPS-sum in the multivariate case.\\
\begin{equation}
\operatorname{CRPS_{\text {sum }}}=\mathbb{E}_{t}\left[\operatorname{CRPS}\left(\widehat{P}_{\text {sum }}(t), \sum_{i} x_{i}^{t}\right)\right]
\end{equation}\textbf{Baselines} We include several baseline methods. For the classical settings and competitive baselines probabilistic models, we follow the comparison methods in \cite{rasul2020multivariate}: VAR-Lasso(a Lasso regularized VAR):  models enable effective selection from a large set of potential explanatory variables of the most relevant regressors for minimizing forecasting errors without complex sampling methods, GARCH \cite{van2002go}: The generalized orthogonal GARCH model is proposed to estimate the time-varying covariance matrix using independent component analysis (ICA) to reduce the parameters to be estimated, Gaussian process model (GP): is a probabilistic modelling approach with a specific property that the mapping between regression inputs and output is presented with a stochastic process and, consequently, the model prediction with a distribution, expressed in terms of mean value and the variance of the modelled variable, KVAE \cite{krishnan2017structured}: disentangles the observations and the latent dynamics (transitions) with VAE, VES \cite{hyndman2008forecasting}, Vec-LSTM-ind-scaling, Vec-LSTM-lowrank-Copula, GP-scaling and GP-Copula. For the combination of normalizing flow and sequence modeling, we take LSTM-Real-NVP, LSTM-MAF and Transformer-MAF as the competitive auto-regressive baselines \cite{rasul2020multivariate}: These models leverage LSTM and Transformer as the encoder module of historical sequences, and introduce the masked autoregressive flow into the inference stage. Moreover, for the non-autoregressive modeling, we select NKF \cite{de2020normalizing}: a novel approach for modeling and forecasting complex multivariate time series by augmenting classical linear Gaussian state space models (LGM) with normalizing flows and HMGT \cite{ding2020hierarchical}: a novel hierarchical Gaussian Transformers for modeling stock prices, instead of relying on recurrent neural networks in our work.\\
\begin{table*}[htb]
  \renewcommand\arraystretch{1.3}
  \centering
  \caption{The Test set CRPS-sum(C-S) and MSE comparison(lower is better) of models from the baselines and our model MANF, with - are runs failed with numerical issues.(*) indicates the experimental variance.}
  \resizebox{1.0\linewidth}{!}{
    \begin{tabular}{lrr|rr|rr|rr|rr|rr}
    \toprule
          & \multicolumn{2}{c}{EXCHANGE} & \multicolumn{2}{c}{SOLAR} & \multicolumn{2}{c}{ELECTRICITY} & \multicolumn{2}{c}{TRAFFIC} & \multicolumn{2}{c}{TAXI} & \multicolumn{2}{c}{WIKIPEDIA} \\
\cmidrule{2-13}    Method & \multicolumn{1}{c}{C-S} & \multicolumn{1}{c|}{MSE} & \multicolumn{1}{c}{C-S} & \multicolumn{1}{c|}{MSE} & \multicolumn{1}{c}{C-S} & \multicolumn{1}{c|}{MSE} & \multicolumn{1}{c}{C-S} & \multicolumn{1}{c|}{MSE} & \multicolumn{1}{c}{C-S} & \multicolumn{1}{c|}{MSE} & \multicolumn{1}{c}{C-S} & \multicolumn{1}{c}{MSE} \\
    \midrule
    VAR-Lasso &  0.012(.001)    &   -    &   0.510(.006)    &   -    &   0.025(.000)     &   -    &  0.150(.002)    &   -    &   -    &    -   &   3.10(.004)    & - \\ \hline
    GP   &  0.011(.001)    &  -    &  0.828(.010)   &  -     &   0.947(.016)     &  -     &   2.198(.774)     &  -    &    0.425(.199)      &   -    &   0.93(.003)     & - \\ \hline
    GARCH &   0.020(.000)    &    -   &  0.869(.000)     &  -     &   0.278(.000)     &   -    &  0.368(.000)      &  -     &   -    &   -    &    -   & - \\ \hline
    VES   &  0.005(.000)   &   -    &  0.900(.003)      &   -    &   0.880(.003)     &   -    &   0.350(.002)     &  -     &    -   &    -   &   -    &  -\\ \hline
    KVAE  &   0.014(.002)   &   -    &   0.340(.025)     &   -    &  0.051(.019)      &   -    &  0.100(.005)      &   -    &   -    &    -   &   0.095(.012)     & - \\ \hline
    \multicolumn{1}{p{5.0em}}{Vec-LSTM\newline{}ind-scaling} & \centering 0.008(.000)   &  \textbf{1.6e-4}     &   0.391(.012)     &  9.3e2 &   0.025(.001)     &   2.1e5  &  0.087(.037)   &   6.3e-4  &  0.506(.004)   &  7.3e   &   0.133(.004)     & 7.2e7 \\ \hline
    \multicolumn{1}{p{5.0em}}{Vec-LSTM\newline{}lowrank-Copula} &  0.017(.000)     &  1.9e-4   &   0.319(.010)     &  2.9e3    &   0.064(.006)      &  5.5e6    &  0.103(.004)   &  1.5e-3     &   0.326(.004)     &  5.1e     &   0.241(.001)      &  3.8e7 \\ \hline
    GP-scaling &  0.009(.000)     &   2.9e-4    &   0.368(.009)     &  1.1e3     &   0.022(.000)     &  1.8e5     &   0.079(.002)     &   5.2e-4    &   0.183(.218)     &   2.7e    &    1.483(1.017)    & 5.5e7  \\ \hline
    GP-Copula &   0.008(.000)   &   1.7e-4    &  0.337(.022)      &  9.8e2     &  0.024(.002)      &  2.4e5     &  0.078(.001)      &  6.9e-4     &  0.208(.201)      &   3.1e    &  0.086(.000)      & 4.0e7 \\ \hline
    \multicolumn{1}{p{5.0em}}{LSTM-\newline{}Real-NVP} & 0.006(.001)      &   2.4e-4    &  0.331(.020)      &  9.1e2     &  0.024(.001)      &     2.5e5  &   0.078(.001)     &  6.9e-4     &   0.175(.001)     &  2.6e     &  0.078(.000)      &  4.7e7 \\ \hline
    LSTM MAF &  0.005(.003)      &  3.8e-4 &  0.315(.032)      &   9.8e2    &  0.020(.000)      &    1.8e5   &   0.069(.002)    &   4.9e-4    &   0.161(.002)     &   2.4e    &   0.067(.002)     &  3.8e7 \\ \hline
    \multicolumn{1}{p{5.0em}}{Transformer\newline{}MAF} &   0.005(.003)     &   3.4e-4   &  0.301(.001)      &  9.3e2     &  0.020(.000)      &  2.0e5     &  0.056(.001)      &  5.0e-4     &   0.179(.002)     &   4.5e    &   0.063(.007)     & 3.1e7 \\ \hline
    HMGT   &   0.005(.001)     &  3.7e-4     &   0.327(.014)     &   9.4e2    &   0.022(.003)     &  2.1e5     &   0.052(.002)     &    4.4e-4   &   0.158(.016)     &   2.4e    &   0.074(.010)     & 3.0e7 \\ \hline
    NKF   &   0.005(.001)     &  -     &   0.320(.007)     &   -    &   0.016(.002)     &  -     &   0.100(.003)     &    -   &   -    &   -    &   0.071(.001)     & - \\ \hline
    \textbf{MANF(ours)} &  \textbf{0.004}(.000)      &  2.8
    e-4     &   \textbf{0.253}(.001)     &  \textbf{7.4e2}    &   \textbf{0.014}(.003)     &   \textbf{1.6e5}    &   \textbf{0.026}(.001)     &  \textbf{4.1e-4}     &  \textbf{0.123}(.002)      &   \textbf{2.2e}    &  \textbf{0.057}(.002)      &  \textbf{2.87e7}\\ 
    \bottomrule
    \end{tabular}%
    }
  \label{tab:result}%
\end{table*}\textbf{Implementation detail} Our method is dependent upon the ADAM optimizer with an initial learning rate of $5 e^{-4}, 1 e^{-3}$, the batch size is 64, the hidden size of the multi-scale and self-attention are 32, the normalizing flow layer is 32, 64 and the training process is 60 epochs. All experiments are repeated three times, implemented in PyTorch \cite{paszke2019pytorch} and GluonTS \cite{alexandrov2020gluonts}. The hyper-parameter of scale set $\Theta$ is $[1/3L, 1/2L, L]$, and L is 4 times the predicted length. Our MANF contains 3 encoder layers and 3 decoder layers (3 conditional normalizing flow layers). Besides, in the ablation studies, MANF-T replaced the 3-layer multi-scale attention mechanism of the encoder with the three Transformer encoder layers, and the number of decoder layers and the corresponding normalizing flow structure is the same as MANF.

Specifically, the batch sizes are set to 64 throughout our training, with 100 batches per epoch, and train for a maximum of 60 epochs with a learning rate of $5 e^{-4}$ (Exchange, WIKIPEDIA), $1 e^{-3}$ (SOLAR, ELECTRICITY, TRAFFIC, TAXI). We used 3 stacks of normalizing flow bijections layers. The components of the normalizing flows are linear feed-forward layers (with fixed input and final output sizes because we model bijections) with hidden dimensions of 100. We sample 100 times to report the metrics on the test set. In the three data sets EXCHANGE, SOLAR, and ELECTRICITY, the multi-scale attention coding layer and self-attention decoding layer all leverage 4 heads, and the number of attention heads in the remaining data set is 8 and a dropout rate of 0.1. All experiments run on Nvidia 2080Ti GPUs and all experiments and ablation studies are repeated more than five times.
\subsection{Main Results}
We compare the test time prediction of our MANF to the above baselines with CRPS, CRPS-sum, and MSE. The results for probabilistic forecasting in a multivariate setting are shown in Tab.~\ref{tab:result}. We observe that MANF achieves the state-of-the-art (to the best of our knowledge) CRPS-sum on almost all benchmarks. Notably, our model has shown a significant MSE reduction in Solar $24  \%(9.8e2 \rightarrow 7.4e2)$, in Electricity $11  \%(1.8e5 \rightarrow 1.6e5)$, in Traffic $17  \%$(4.9e-4 $\rightarrow$ 4.1e-4) and $23  \%(3.8e7 \rightarrow 2.87e7)$ in Wikipedia. Note that our model provides mediocre performance in the Exchange dataset that is without obvious periodicity. However, it is not difficult to find that in Solar and Traffic datasets with relatively periodic and stable, MANF greatly exceeds the other baselines, which shows that the multi-scale attention mechanism with dynamic relative position embedding can effectively capture local contextual information and long-period patterns. Moreover, in Taxi (1214 dimensions) with high stochasticity, MANF also shows excellent results, indicating that in high-dimensional multivariate time series, it is important to model the true data distribution with tractable likelihoods rather than some approximately generative model. To verify the effectiveness of the non-autoregressive structure, we introduce noise into the encoding sequence and increase the prediction length. The details are shown in Table~\ref{tab:addlabel}.
\begin{figure*}[htp]
\begin{center}
   \includegraphics[width=0.75\textwidth]{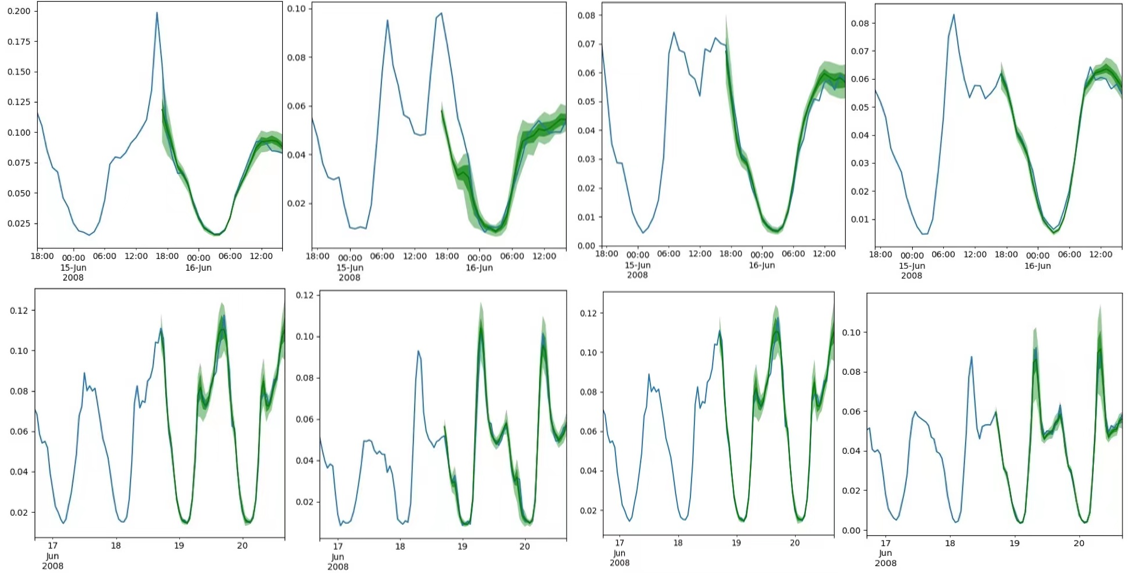}
  \caption{MANF one and twice prediction intervals and test set ground-truth for Traffic data of the 8 of 963 dimensions from a rolling window.}
  \label{okk}
\end{center}
\end{figure*}

\begin{figure*}[htp]
\begin{center}
   \includegraphics[width=0.75\textwidth]{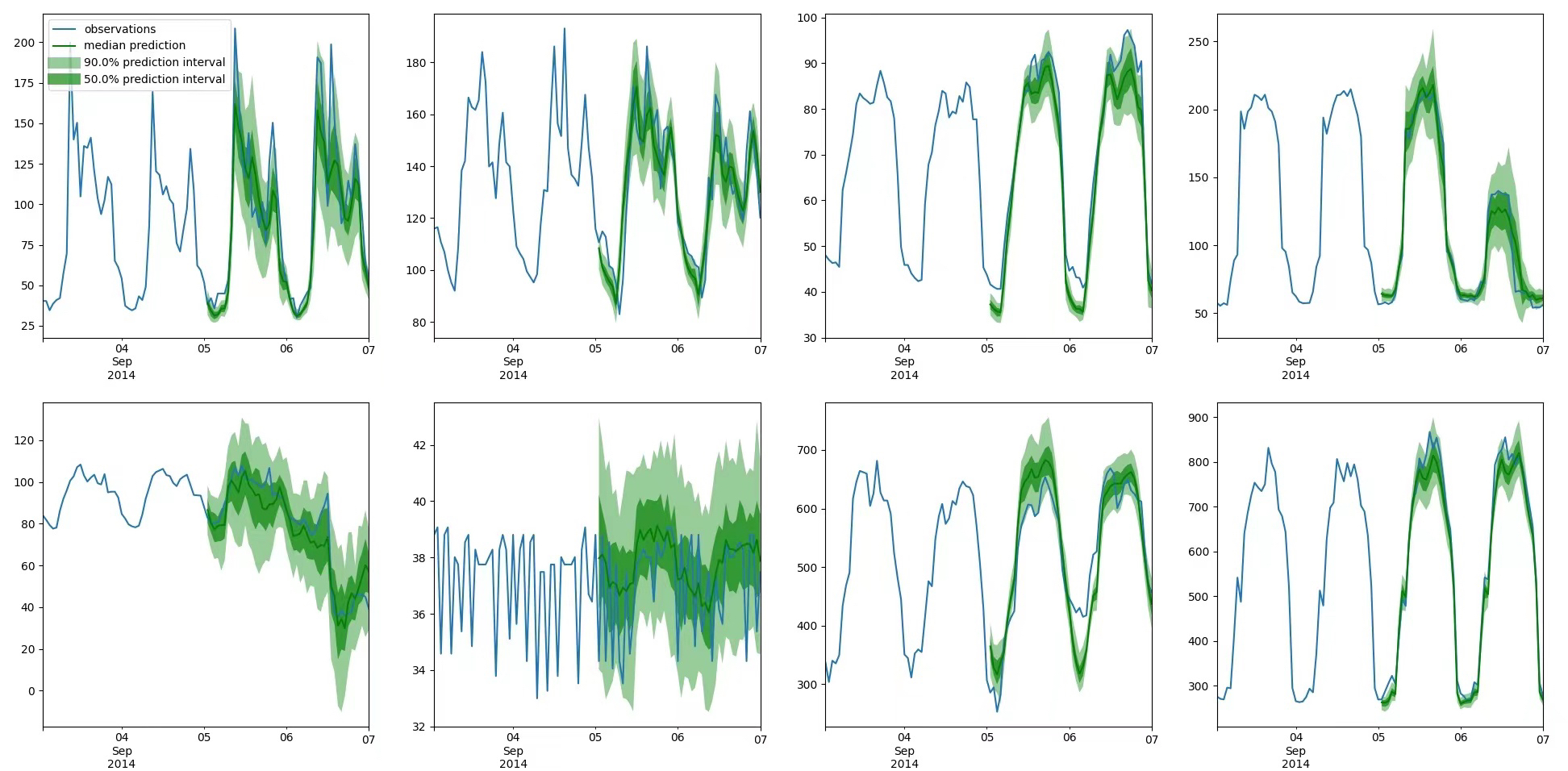}
  \caption{MANF twice prediction intervals and test set ground-truth for Electricity data of the 8 of 370 dimensions from a rolling window.}
  \label{okk}
\end{center}
\end{figure*}

Moreover, to highlight the predictions of our MANF we show in Fig.~\ref{okk} and Fig.~\ref{okk_2} the predicted median, 50th percentile, and 90th percentile distribution intervals of 8 dimensions of the full 963-dimensional multivariate forecast of the traffic, electricity, and taxi benchmark (The predicted length includes 24 and 48). Further experimental results are reported in the appendix. We observed that our model appears to capture the uncertainty of future forecasts to some extent; observations of large magnitudes and far into the future seem to correctly have higher variance estimates, indicating that if a powerful longer sequence modeling method with this non-autoregressive way as the backbone into our normalizing flows, the impact caused by the passage of time will be reduced. However, as the prediction sequence grows, our MANF will not be significantly affected. Within Table~\ref{tab:addlabel}, we conducted a comparative experiment about increasing the prediction length in three datasets of different magnitudes. It is not difficult to find that, compared with the autoregressive structure, the performance of MANF has weak even with no attenuation which proves that our one-shot MANF can resist the negative impact of the increase of the prediction length.

Specifically, under twice the predicted length interval condition, the performance of two autoregressive flow models on three datasets is lower than our MANF. Among them, we found that with the increasing dimensions of the dataset, the advantages of our MANF are more obvious. For example, under the CRPS-sum evaluation standard, the performance of MANF is only improved by 25\% on the Solar dataset, (0.314 (ours) < 0.354 (+12\%) < 0.392 (24.8\%)). In Traffic (963 dimensions), MANF shows significant performance improvement (0.022 (ours) < 0.085 < 0.617). Also, under the MSE, MANF shows obvious improvement in the three datasets.

\begin{table}[htp] \normalsize %
  \centering
  \caption{C1:twice the predicted length, C2: 30\% missing value, C3: 50\% missing value. Here, we add different degrees of missing values to improve the noise level of the sequences. The Test set CRPS-sum(C-S) and MSE comparison(lower is better) of models including Transformer-MAF, LSTM-MAF and our model MANF. (*) indicates the experimental variance.}
    \resizebox{1.0\linewidth}{!}{ \begin{tabular}{c|c|cccccc}
    \toprule
    \multicolumn{2}{c}{Models} & \multicolumn{2}{c}{\textbf{MANF}} & \multicolumn{2}{c}{Trans-MAF} & \multicolumn{2}{c}{LSTM-MAF} \\
\cmidrule{3-8}    \multicolumn{2}{c}{Metric} & C-S   & MSE   & C-S   & MSE   & C-S   & MSE \\
    \midrule
    \multirow{3}[1]{*}{\begin{sideways}Solar\end{sideways}} & C1    & \textbf{0.314}(.002) & \textbf{7.9e2} & 0.354(.001) & 9.8e2 & 0.392(.001) & 1.1e3 \\
          & C2    & \textbf{0.254}(.001) & \textbf{7.4e2} & 0.368(.002) & 9.9e2     & 0.396(.002) & 1.2e3 \\
          & C3    & \textbf{0.283}(.002) & \textbf{7.6e2} & 0.412(.001) & 1.3e3    & 0.674(.004) & 2.6e3 \\ 
    \midrule
    \multirow{3}[1]{*}{\begin{sideways}\footnotesize{Electricity}\end{sideways}} & C1    & \textbf{0.023}(.001) & \textbf{1.8e5} & 0.028(.000) & 2.3e5   & 0.029(.003) & 2.7e5 \\
          & C2    & \textbf{0.020}(.001) & \textbf{1.6e5} & 0.037(.001) & 2.9e5   & 0.042(.002) & 3.4e5 \\
          & C3    & \textbf{0.024}(.000) & \textbf{1.8e5} & 0.044(.001) & 3.1e5   & 0.105(.003) & 6.7e5 \\
    \midrule
    \multirow{3}[1]{*}{\begin{sideways}Traffic\end{sideways}} & C1    & \textbf{0.022}(.001) & \textbf{3.9e-4} & 0.085(.000) & 6.3e-4   & 0.617(.002) & 4.2e-3 \\
          & C2    & \textbf{0.026}(.001) & \textbf{4.1e-4} & 0.082(.001) & 6.1e-4   & 0.919(.003) & 8.7e-3 \\
          & C3    & \textbf{0.039}(.000) & \textbf{4.4e-4} & 0.094(.001) & 6.6e-4   & 1.724(.002) & 5.4e-2 \\
    \bottomrule
    \end{tabular}%
    }
  \label{tab:addlabel}%
\end{table}%


\begin{figure*}[ht]
\begin{center}
  \includegraphics[width=0.75\textwidth]{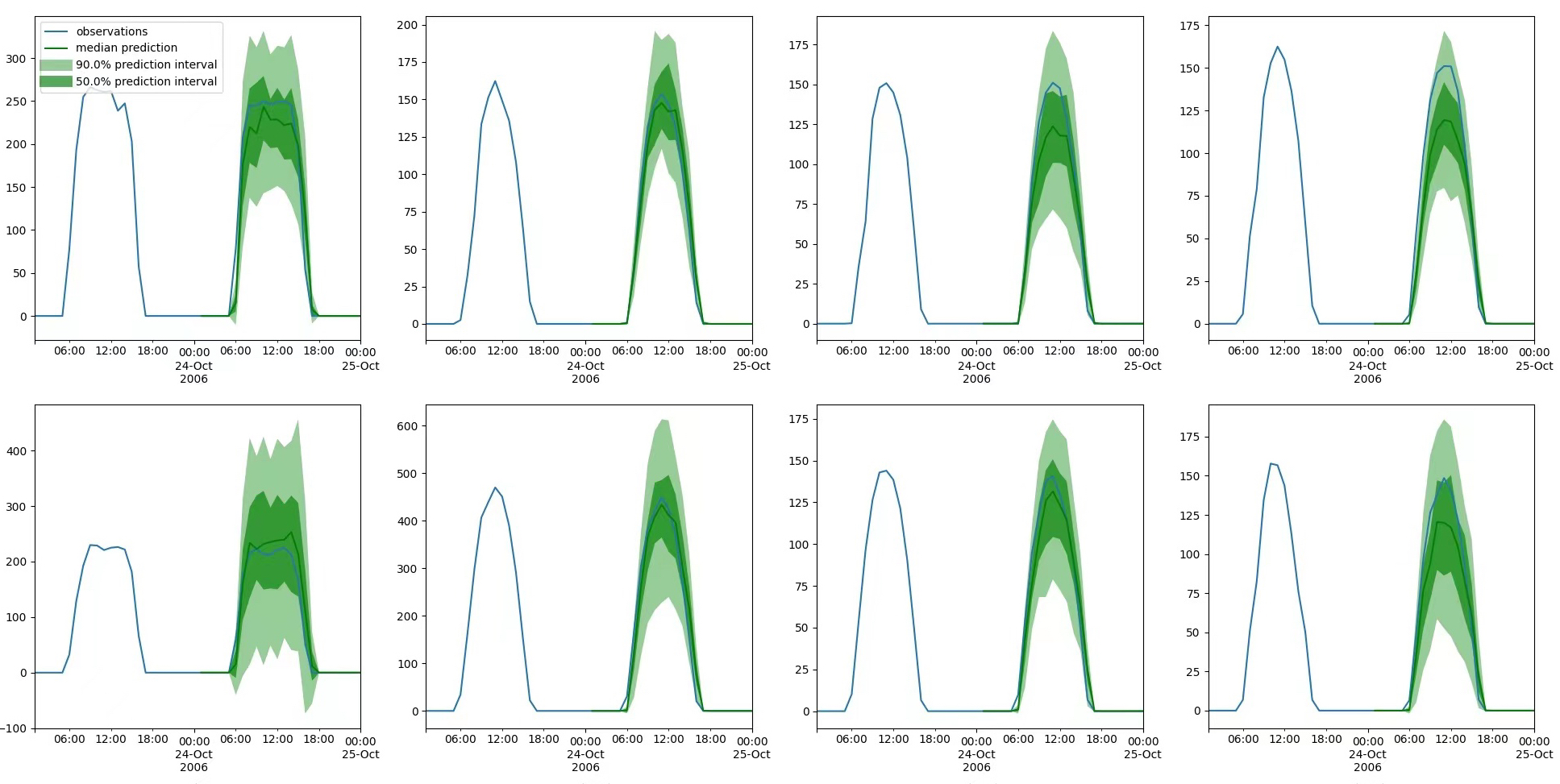}
  \caption{MANF prediction intervals and test set ground-truth for Solar data of the 8 of 137 dimensions from a rolling window}
  \label{solar_main}
\end{center}
\end{figure*}

\begin{figure*}[ht]
\begin{center}
  \includegraphics[width=0.75\textwidth]{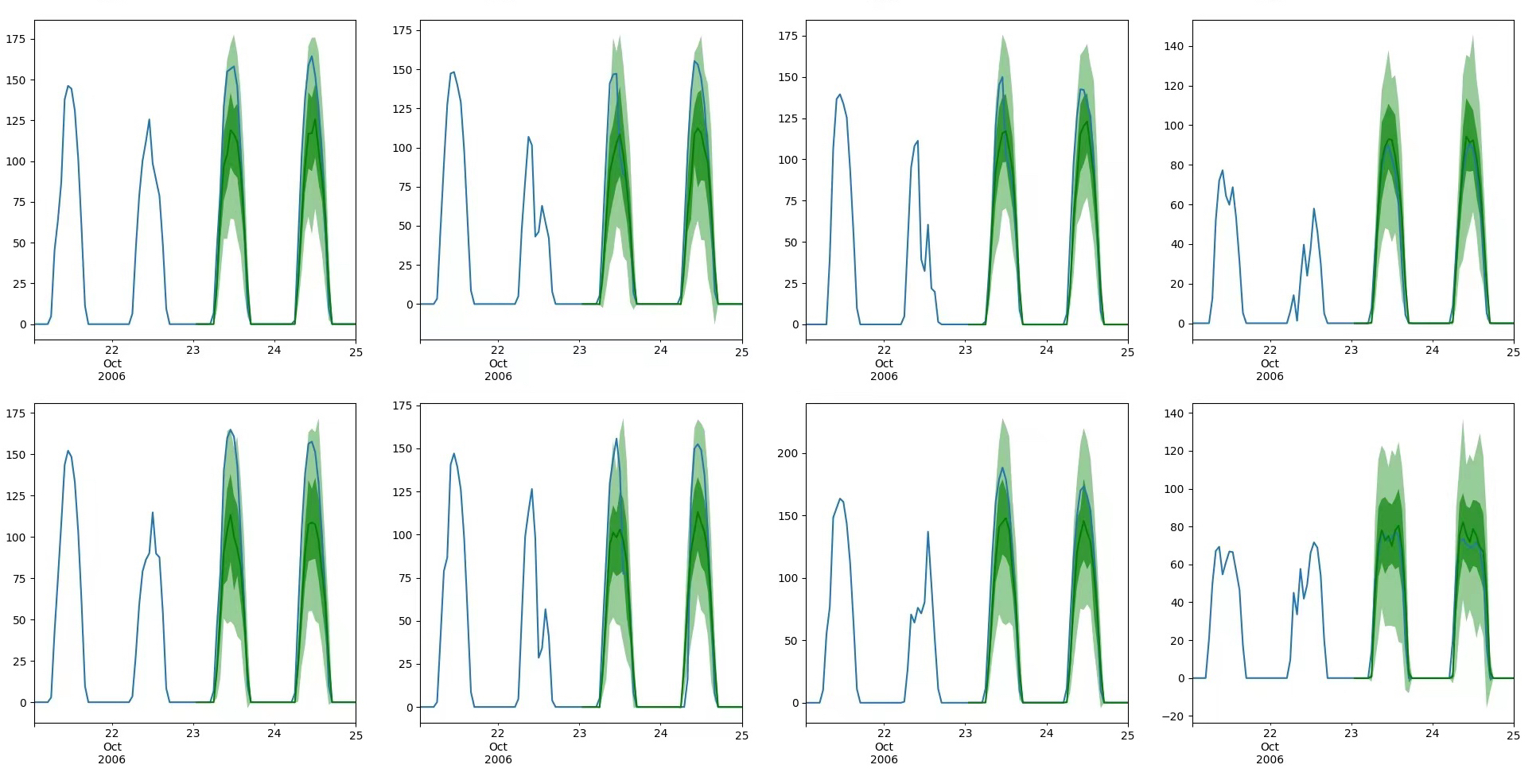}
  \caption{MANF twice prediction intervals and test set ground-truth for Solar data of the 8 of 137 dimensions from a rolling window}
  \label{solar_main}
\end{center}
\end{figure*}


\subsection{Ablation studies}
With our proposed architecture being a hierarchical modeling method, we take the output of the last layer in our decoder as the input to the stacked normalizing flows (MANF-L) or utilize the vanilla Transformer as sequence modeling structure (MANF-T). This verifies that our generative model can study the hierarchical information within the multivariate data and effectively conduct inductive bias. The experimental results between our model and these methods are shown in Table ~\ref{tab:rebuttal1}

Moreover, our proposed MANF contains two important components, (i) relative position encoding and (ii) multi-scale attention mechanism. Aiming at the influence of multi-scale attention mechanism and relative position on our model performance, we conducted the following experiments, as presented in Table~\ref{tab:rebuttal1}. The experimental results show that the performance of MANF (ours) is greater than that of MANF-P (multi-scale attention with vanilla position embedding) and MANF-T (self-attention replaced multi-scale attention as encoder), indicating that the combination of multi-scale attention mechanism and relative position can effectively improve the performance of our model.


\begin{table*}[ht]
  \renewcommand\arraystretch{1.2}
  \centering
  \caption{The Test set CRPS-sum(C-S) and MSE comparison(lower is better) of models MANF-T, MANF-P, MANF-M, MANF-L and MANF in Ablation studies. (*) indicates the experimental variance.}
  \resizebox{1.0\linewidth}{!}{ 
    \begin{tabular}{lrrrrrrrrrrrr}
    \toprule
          & \multicolumn{2}{c}{Exchange} & \multicolumn{2}{c}{Solar} & \multicolumn{2}{c}{Electricity} & \multicolumn{2}{c}{Traffic} & \multicolumn{2}{c}{Taxi} & \multicolumn{2}{c}{Wikipedia} \\
\cmidrule{2-13}    Method & \multicolumn{1}{c}{C-S} & \multicolumn{1}{c}{MSE} & \multicolumn{1}{c}{C-S} & \multicolumn{1}{c}{MSE} & \multicolumn{1}{c}{C-S} & \multicolumn{1}{c}{MSE} & \multicolumn{1}{c}{C-S} & \multicolumn{1}{c}{MSE} & \multicolumn{1}{c}{C-S} & \multicolumn{1}{c}{MSE} & \multicolumn{1}{c}{C-S} & \multicolumn{1}{c}{MSE} \\
    \midrule
    MANF-P &   0.005(.001)     &   3.2e-4    &  0.277(.007)      &  7.5e2     &  0.021(.001)      &  1.7e5    &  0.031(.002)      &  4.2e-4     &   0.134(.002)     &   2.6e    &   0.067(.001)     & 3.24e7 \\ \hline
    MANF-M &   0.005(.002)     &   \textbf{2.7e-4}    &  0.255(.003)      &  7.7e2     &  \textbf{0.014}(.002)      &  \textbf{1.6e5}    &  \textbf{0.025}(.001)      &  \textbf{4.0e-4}     &   0.125(.001)     &   \textbf{2.2e}    &   0.058(.002)     & 2.91e7 \\ \hline
   MANF-T   &   0.009(.006)     &   4.8e-4    &   0.500(.010)     &   1.2e3    &   0.034(.002)     &  3.1e5    &   0.137(.004)     &    7.1e-4  &   0.176(.000)     &   2.4e    &   0.082(.003)     &4.00e7 \\ \hline
     MANF-L &   0.005(.001)     &   3.0e-4    &  0.303(.002)      &  7.7e2     &  \textbf{0.014}(.002)      &  1.8e5    &  0.028(.002)      &  4.3e-4     &   0.128(.001)     &   2.3e    &   0.061(.002)     & 3.10e7 \\ \hline
    \textbf{MANF(ours)} &  \textbf{0.004}(.000)      &  2.8e-4     &   \textbf{0.253}(.001)     &  \textbf{7.4e2}    &   \textbf{0.014}(.003)     &   \textbf{1.6e5}    &   0.026(.001)     &  4.1e-4     &  \textbf{0.123}(.002)      &   \textbf{2.2e}    &  \textbf{0.057}(.002)      &  \textbf{2.87e7}\\ 
    \bottomrule
    \end{tabular}%
    }
  \label{tab:rebuttal1}%
\end{table*}

We also notice that the result of MANF-P is lower than that of MANF but much higher than that of MANF-T. The reason for this result is that the position information and content feature of every time point are learned separately in formula 6 of our manuscript, and the wrong location will lead to a bad effect on model performance. Therefore, the multi-scale attention mechanism and relative position should be a complete whole.


We observed that the effect of MANF-T has a large deviation, indicating self-attention can not capture the accurate correlation between time sequences. Compared with MANF, MANF-L shows better performance than most baselines, such as $50  \%(0.056 \rightarrow 0.028)$ reduction in CRPS-sum on Traffic, $21  \%(0.161 \rightarrow 0.128)$ on Taxi, $9  \%(0.320 \rightarrow 0.303)$ on Solar dataset.


Finally, In the decoder of MANF, instead of leveraging the same multi-scale attention as the encoder, we choose the vanilla Transformer decoder. Therefore, we replaced the self-attention of the decoder with multi-scale attention (MANF-M), and the experimental results (CRPS-Sum/MSE) are described in Table ~\ref{tab:rebuttal1}. According to the experimental results, it is not difficult to see that the multi-scale attention of the decoder (MANF-M) is slightly different from that of MANF. But in our ablation studies, MANF-T converts multi-scale attention to self-attention in our encoder, and we find that the performance of the model is significantly degraded.\\
\begin{figure*}[ht]
\begin{center}
\centerline{\includegraphics[width=1.0\textwidth]{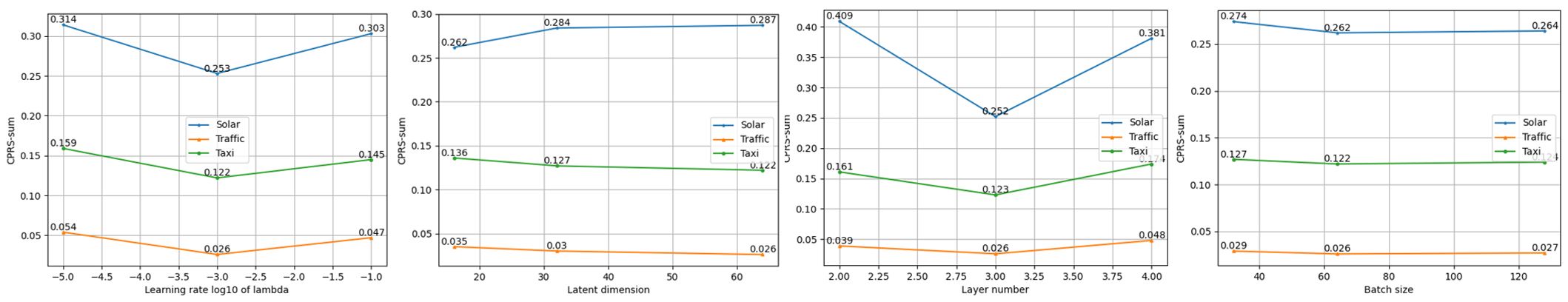}}
\caption{The experimental results(CRPS-sum) of three datasets with significant differences in dimensions within the varying batch size, learning rate, layer number, and latent dimension.}
\label{sensitive}
\end{center}
\end{figure*}\textbf{Parameter sensitivity study} In order to further illustrate the effectiveness of our model and explore more excellent forecasting settings, we analyzed four hyperparameters: batch size, learning rate, the number of the encoder/decoder layer and the latent dimension. Figure ~\ref{sensitive} shows all the experimental results on parameters.

Fig ~\ref{sensitive} illustrates the metric variability with an increasing learning rate. As for the performance of CRPS-sum, we found that with the continuous increase in the learning rate, the three datasets with different orders of magnitude have the same trend of variety, that is, the V-shaped fluctuation, but the amplitude is tiny, which illustrates that MANF can effectively fit on different 
dimensions of datasets without large performance fluctuations. Besides, in the parameter experiment of the encoder and decoder layers, we also got a similar V-shaped fluctuation. However, on the Solar dataset, the performance of the model fluctuated significantly, and the other changes were not obvious. We believe that the combination of the deep networks and the minor dimensional dataset will cause the model to get into the fitting state, so there will be a large fluctuation. Therefore, the model performance slightly improves to a relatively stable case with the increase in the layer number, and learning rate, indicating higher layer number and learning rate may not help considerably. 

Next, in the experimental results on latent dimensions, the solar dataset has a different shape, that is, as the potential dimensions continue to increase, its performance continues to decline. This explanation is the same as the above deep network of the model. Solar could show excellent performance in moderate network structure and small latent dimensions.

Lastly, we vary the different batch sizes and found that it has no great influence on the performance of our model.\\ 
\begin{table}[t]
  \renewcommand\arraystretch{1.1}
  \centering
  \caption{Average training time per epoch and average testing time of MANF and other baselines on Electricity}
  \resizebox{0.95\columnwidth}{!}{ 
    \begin{tabular}{lll}
    \toprule
     Method & Training & Testing  \\ \hline
 LSTM MAF & 31.27 $\pm$ 0.003 & 34.19 $\pm$ 0.002  \\ \hline
 Transformer MAF &  7.250 $\pm$ 0.002    &   14.36 $\pm$ 0.002 \\ \hline
    \textbf{MANF}   &  \textbf{2.179 $\pm$ 0.001}   &   \textbf{2.192 $\pm$ 0.001} \\ 
    \bottomrule
    \end{tabular}
     }
  \label{tab:time}%
\end{table}\textbf{Efficiency and scalability analysis}
Firstly, in terms of efficiency, compared with the existing autoregressive models, our non-autoregressive MANF performs nicely. The one-shot normalizing flow structure can considerably improve the training and prediction efficiency, which could be drawn from Table ~\ref{tab:time}. Moreover, to illustrate the scalability of MANF, we tested it under three different stress conditions (e.g. twice the predicted length, 30\% missing value and 50\% missing value). Within Table ~\ref{tab:addlabel}, it is obviously seen that the performance of the autoregressive flow models declines exponentially with the increase of the prediction length and missing values, which is evidently inferior to our MANF. This kind of stress test more efficiently illustrates the scalability of MANF. 

Specifically, we found that, with the increasing ratio of missing values, there would be no significant performance degradation in MANF on tiny dimension datasets. In the large dimension dataset like traffic, when the environment is C3 (50\% missing value), MANF has a certain degree of performance degradation on the CRPS-sum. However, compared with other flow models, MANF still far exceeds the comparative structure. For example, under the MSE evaluation standard, MANF has obvious advantages, (7.9e2 (ours) < 9.8e2 (+24\%) < 1.1e3 (+39.2\%)) in Solar ; (1.8e5 (ours) < 2.3e5 (+27\%) < 2.7e5 (+50\%)) in Electricity ; (3.9e-4 (ours) < 6.3e-4 (+61.5\%) < 4.2e-3) in Traffic. In summary, compared with the autoregressive method, MANF not only improves the accuracy of prediction but also considerably boosts efficiency and scalability.

\section{Conclusion}
We have presented MANF, a versatile multivariate probabilistic time series forecasting method that leverages multi-scale attention to study the hierarchical information of the time series sequence. Dependent upon the normalizing flows, we generate the multivariate distribution in a non-autoregressive way. Analysis of MANF on six commonly used time-series benchmarks establishes the new state-of-the-art against competitive methods. Moreover, our one-shot MANF avoids the negative impact of accumulated errors and the dilemma of long reasoning time compared with the autoregressive model, and our multi-layer normalizing flow remedies the possible accuracy problems with non-autoregressive models. In summary, our non-autoregressive generative model increases the possibility of applying it to some realistic scenarios.

A natural way to improve our method is some improvements to flows, e.g. FLow++ \cite{ho2019flow++} to model the multivariate data distribution. Since the strong stochasticity of some generated sequences (e.g. Taxi) requires extremely nonlinear normalizing flow, which needs to be completed by stacking multilayers. However, multi-layer decoders may result in large deviations in the generated results. In the future, we will explore more powerful nonlinear flow to correspond to our non-autoregressive model, and explore some medical fields to apply our model. To our knowledge, the use of normalizing flows for discrete-valued data dictates that one dequantizes it. However, dequantization is not used in our setting and future work could explore methods of explicitly modeling the discrete distributions (e.g. sales data). In the future, we will further explore the combination of generative models and sequence models, and continue to improve the multivariate probabilistic time series forecasting method for better effectiveness, stability, and scalability. For example, we can further explore the non-autoregressive based denoising diffusion probabilistic models\cite{ho2020denoising} to accomplish the forecasting task. 

\ifCLASSOPTIONcaptionsoff
  \newpage
\fi

\end{document}